\documentclass[11pt,a4paper]{article}
\usepackage[hyperref]{naaclhlt2019}
\usepackage{times}
\usepackage{latexsym}

\usepackage{url}

\usepackage{epsfig}
\usepackage{graphicx}
\usepackage{xcolor}
\usepackage{footnote}
\usepackage{color}
\usepackage{amsmath}
\usepackage{amssymb}
\usepackage{multirow, varwidth}
\usepackage{dblfloatfix}
\usepackage{verbatim}
\makeatletter
\newif\if@restonecol
\makeatother

\usepackage[ruled,vlined]{algorithm2e}
\usepackage{xr}
\usepackage[amsmath,thmmarks]{ntheorem}

\DeclareRobustCommand\onedot{\futurelet\@let@token\@onedot}
\def\onedot{. }
\def\eg{\emph{e.g}\onedot} 
\def\ie{\emph{i.e}\onedot}

\newcommand{\new}[1]{#1}
\usepackage{tabu}

\aclfinalcopy 
\def\aclpaperid{131} 

\title{Abstractive Summarization of \textit{Reddit} Posts \\with Multi-level Memory Networks}

\author{
    Byeongchang Kim \qquad Hyunwoo Kim \qquad Gunhee Kim \\
    Department of Computer Science and Engineering \& Center for Superintelligence \\
    Seoul National University, Seoul, Korea \\
    {\tt\small \{byeongchang.kim,hyunwoo.kim\}@vision.snu.ac.kr gunhee@snu.ac.kr} \\ 
    \url{http://vision.snu.ac.kr/projects/reddit-tifu}
}

\date{}

\begin{document}
\maketitle

\begin{abstract}
We address the problem of abstractive summarization in two directions: proposing a novel dataset and a new model.
First, we collect \textit{Reddit TIFU} dataset, consisting of 120K posts from the online discussion forum Reddit.
We use such informal crowd-generated posts as text source, in contrast with existing datasets that mostly use formal documents as source such as news articles.
Thus, our dataset could less suffer from some biases that key sentences usually locate at the beginning of the text and favorable summary candidates are already inside the text in similar forms. 
Second, we propose a novel abstractive summarization model named \textit{multi-level memory networks} (MMN), 
equipped with multi-level memory to store the information of text from different levels of abstraction.
With quantitative evaluation and user studies via Amazon Mechanical Turk, we show the \textit{Reddit TIFU} dataset is highly abstractive and the MMN outperforms the state-of-the-art summarization models.
\end{abstract}

\section{Introduction}
\label{sec:introduction}

Abstractive summarization methods have been under intensive study,
yet they often suffer from inferior performance compared to extractive methods~\cite{Allahyari:2017:arXiv, Nallapati:2017:AAAI, See:2017:ACL}.
Admittedly, by task definition, abstractive summarization is more challenging than extractive summarization.
However, we argue that such inferior performance is partly due to some biases of existing summarization datasets.
The source text of most datasets~\cite{Over:2007:IPM, Hermann:2015:NIPS, Cohan:2018:NAACL-HLT, Grusky:2018:NAACL-HLT, Narayan:2018:EMNLP} originates from formal documents such as news articles,
which have some structural patterns of which extractive methods better take advantage.

In formal documents, there could be a strong tendency that key sentences locate at the beginning of the text 
and favorable summary candidates are already inside the text in similar forms.
Hence, summarization methods could generate good summaries by simply memorizing keywords or phrases from particular locations of the text.
Moreover, if abstractive methods are trained on these datasets, they may not show much abstraction \cite{See:2017:ACL},
because they are implicitly forced to learn structural patterns \cite{Kedzie:2018:EMNLP}.
\citet{Grusky:2018:NAACL-HLT} and \citet{Narayan:2018:EMNLP} recently report similar extractive bias in existing datasets.
They alleviate this bias by collecting articles from diverse news publications or regarding intro sentences as gold summary.

Different from previous approaches, we propose to alleviate such bias issue by changing the source of summarization dataset.
We exploit user-generated posts from the online discussion forum \texttt{Reddit}, especially \texttt{TIFU} subreddit, which are more casual and conversational than news articles. 
We observe that the source text in \texttt{Reddit} does not follow strict formatting and disallows models to simply rely on locational biases for summarization.
Moreover, the passages rarely contain sentences that are nearly identical to the gold summary.
Our new large-scale dataset for abstractive summarization named as \textit{Reddit TIFU}
contains 122,933 pairs of an online post as source text and its corresponding long or short summary sentence.
These posts are written by many different users, but each pair of post and summary is created by the same user.

Another key contribution of this work is to propose a novel memory network model named \textit{multi-level memory networks} (MMN).
Our model is equipped with multi-level memory networks, storing the information of source text from different levels of abstraction (\ie word-level, sentence-level, paragraph-level and document-level).
This design is motivated by that abstractive summarization is highly challenging and requires not only to understand the whole document, but also to find salient words, phrases and sentences.
Our model can sequentially read such multiple levels of information to generate a good summary sentence.

Most abstractive summarization methods \cite{See:2017:ACL,Li:2017:EMNLP,Zhou:2017:ACL,Liu:2018:ICLR,Cohan:2018:NAACL-HLT,Paulus:2018:ICLR}
employ sequence-to-sequence (seq2seq) models \cite{Sutskever:2014:NIPS} where an RNN encoder embeds an input document  and another RNN decodes a summary sentence.
Our MMN has two major advantages over seq2seq-based models.
First, RNNs accumulate information in a few fixed-length memories at every step regardless of the length of an input sequence, and thus may fail to utilize far-distant information due to vanishing gradient. 
It is more critical in summarization tasks, since input text is usually very long ($>$300 words).
On the other hand, our convolutional memory explicitly captures long-term information. 
Second, RNNs cannot build representations of different ranges, since hidden states are sequentially connected over the whole sequence.
This still holds even with hierarchical RNNs that can learn multiple levels of representation. 
In contrast, our model exploits a set of convolution operations with different receptive fields; hence,
it can build representations of not only multiple levels but also multiple ranges (\eg sentences, paragraphs, and the whole document).
Our experimental results show that  the proposed MMN model improves abstractive summarization performance on both our new Reddit TIFU and existing Newsroom-Abs~\citep{Grusky:2018:NAACL-HLT} and XSum~\citep{Narayan:2018:EMNLP} datasets.
It outperforms several state-of-the-art abstractive models with seq2seq architecture such as \cite{See:2017:ACL, Zhou:2017:ACL, Li:2017:EMNLP}.
We evaluate with quantitative language metrics (\eg perplexity and ROUGE \cite{Lin:2004:TSBO}) and user studies via Amazon Mechanical Turk (AMT).

The contributions of this work are as follows.
\begin{enumerate}
    \vspace{-4pt}\item {
            We newly collect a large-scale abstractive summarization dataset named \textit{Reddit TIFU}.
            As far as we know, our work is the first to use non-formal text for abstractive summarization.
    }
    \vspace{-4pt}\item We propose a novel model named \textit{multi-level memory networks} (MMN).
        To the best of our knowledge, our model is the first attempt to leverage memory networks for the abstractive summarization.
        We discuss the unique updates of the MMN over existing memory networks in Section \ref{sec:related_work}.
        \vspace{-4pt}\item With quantitative evaluation and user studies via AMT, we show that our model outperforms state-of-the-art abstractive summarization methods on both Reddit TIFU, Newsroom abstractive subset and XSum dataset.
\end{enumerate}

\section{Related Work}
\label{sec:related_work}

Our work can be uniquely positioned in the context of the following three topics. 

\textbf{Neural Abstractive Summarization}.
Many deep neural network models have been proposed for abstractive summarization. 
One of the most dominant architectures is to employ RNN-based seq2seq models with attention mechanism 
such as \citep{Rush:2015:EMNLP, Chopra:2016:NAACL-HLT, Nallapati:2016:CoNLL, Cohan:2018:NAACL-HLT, Hsu:2018:ACL, Gehrmann:2018:EMNLP}.
In addition, recent advances in deep network research have been promptly adopted for improving abstractive summarization.
Some notable examples include the use of variational autoencoders (VAEs)~\citep{Miao:2016:EMNLP,Li:2017:EMNLP}, 
graph-based attention~\citep{Tan:2017:ACL},
pointer-generator models \citep{See:2017:ACL}, self-attention networks~\citep{Liu:2018:ICLR}, reinforcement learning~\citep{Paulus:2018:ICLR, Pasunuru:2018:NAACL-HLT},
contextual agent attention~\citep{Celikyilmaz:2018:NAACL-HLT} and integration with extractive models~\citep{Hsu:2018:ACL, Gehrmann:2018:EMNLP}.

Compared to existing neural  methods of abstractive summarization, 
our approach is novel to replace an RNN-based encoder with explicit multi-level convolutional memory.
While RNN-based encoders always consider the whole sequence to represent each hidden state, 
our multi-level memory network exploits convolutions to control the extent of representation in multiple levels of sentences, paragraphs, and the whole text.

\textbf{Summarization Datasets}.
Most existing summarization datasets use formal documents as source text.
News articles are exploited the most, including in DUC~\cite{Over:2007:IPM}, Gigaword~\citep{Napoles:2012:NAACL-HLT},
CNN/DailyMail~\citep{Nallapati:2016:CoNLL,Hermann:2015:NIPS}, Newsroom~\citep{Grusky:2018:NAACL-HLT} and XSum~\citep{Narayan:2018:EMNLP} datasets. 
\citet{Cohan:2018:NAACL-HLT} introduce datasets of academic papers from arXiv and PubMed.
\citet{Hu:2015:EMNLP} propose the LCSTS dataset as a collection of Chinese microblog's short text each paired with a summary.
However, it selects only formal text posted by verified organizations such as news agencies or government institutions.
Compared to previous summarization datasets,
our dataset is novel in that it consists of posts from the online forum Reddit.

Rotten Tomatoes and Idebate dataset \citep{Wang:2016:NAACL-HLT} use online text as source,
but they are relatively small in scale: 3.7K posts of RottenTomatoes compared to 80K posts of TIFU-short as shown in Table \ref{tab:dataset}.
Moreover, Rotten Tomatoes use multiple movie reviews written by different users as single source text, and one-sentence consensus made by another professional editor as summary. 
Thus, each pair of this dataset could be less coherent than that of our TIFU, which is written by the same user. 
The Idebate dataset is collected from short arguments of debates on controversial topics, and thus the text is rather formal.
On the other hand, our dataset contains the posts of interesting stories happened in daily life, and thus the text is more unstructured and informal.

\begin{figure}[t] \begin{center}
    \includegraphics[width=\linewidth]{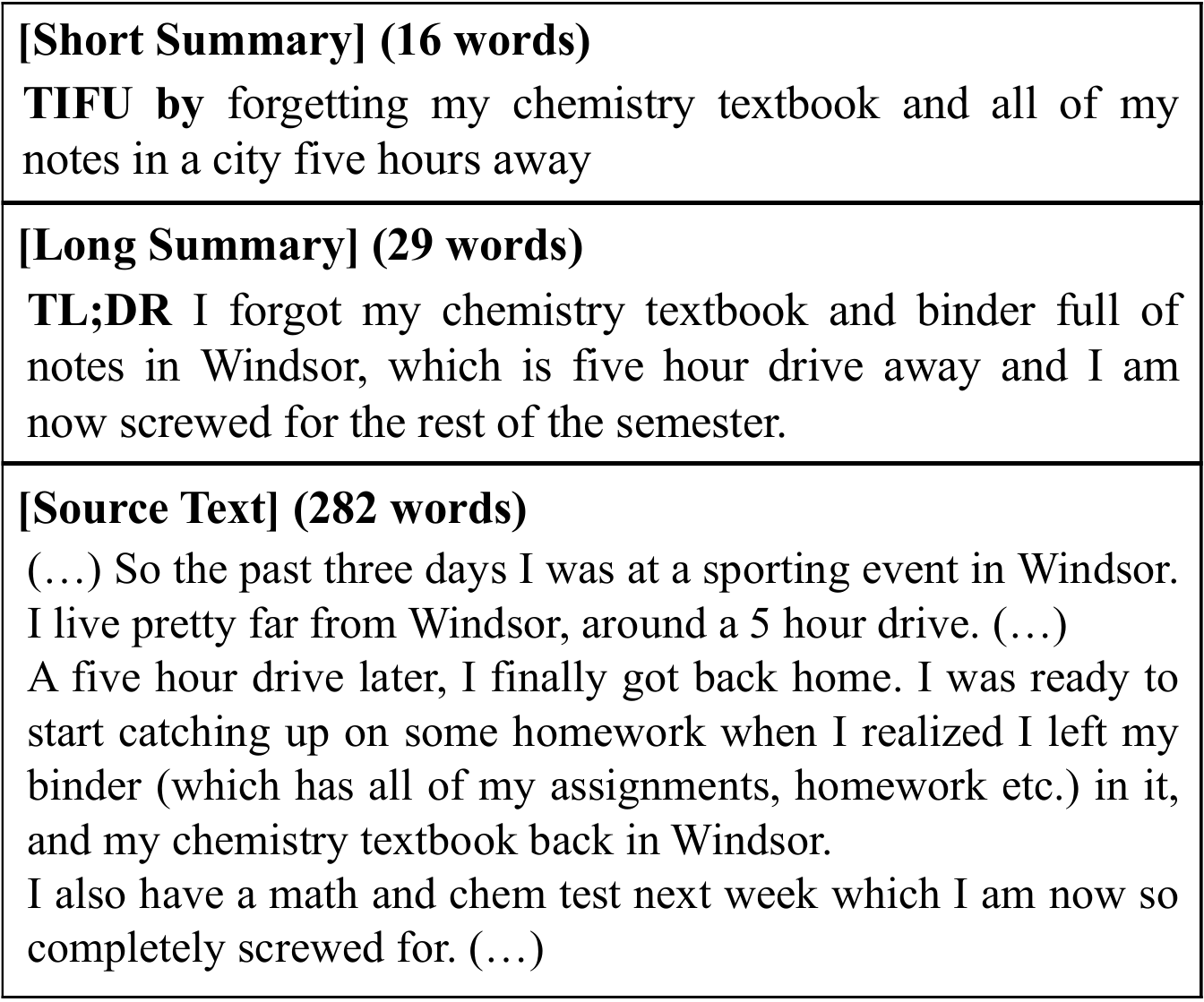}
    \caption{An example post of the \texttt{TIFU} subreddit.}
    \label{fig:post_example}
    \vspace{-5pt}
\end{center} \end{figure}

\textbf{Neural Memory Networks}.
Many memory network models have been proposed to improve memorization capability of neural networks \cite{Kaiser:2017:ICLR, Na:2017:ICCV, Yoo:2019:CVPR}.
\citet{Weston:2014:ICLR} propose one of early  memory networks for language question answering (QA);
since then, many memory networks have been proposed for QA tasks \cite{Sukhbaatar:2015:NIPS,Kumar:2016:ICML,Miller:2016:EMNLP}. 
\citet{Park:2017:CVPR} propose a convolutional read memory network for personalized image captioning.
One of the closest works to ours may be \citet{Singh:2017:CIKM}, which use a memory network for text summarization.
However, they only deal with extractive summarization by storing embeddings of individual sentences into memory.

Compared to previous memory networks, our MMN has four novel features: 
(i) building a multi-level memory network that better abstracts multi-level representation of a long document,
(ii) employing a dilated convolutional memory write mechanism to correlate adjacent memory cells,
(iii) proposing normalized gated tanh units to avoid covariate shift within the network,
and (iv) generating an output sequence without RNNs.

\section{Reddit TIFU Dataset}
\label{sec:reddit_datasets}

We introduce the Reddit TIFU dataset whose key statistics are outlined in Table \ref{tab:dataset}.
We collect data from Reddit, which is a discussion forum platform with a large number of subreddits on diverse topics and interests. 
Specifically, we crawl all the posts from 2013-Jan to 2018-Mar in the \texttt{TIFU} subreddit, 
where every post should strictly follow the posting rules, otherwise they are removed.
Thanks to the following rules\footnote{\url{https://reddit.com/r/tifu/wiki/rules}.}, the posts in this subreddit can be an  excellent corpus for abstractive summarization: 
\textit{Rule 3: Posts and titles without context will be removed. Your title must make an attempt to encapsulate the nature of your f***up.
Rule 11: All posts must end with a TL;DR summary that is descriptive of your f***up and its consequences}.
Thus, we regard the body text as source, the title as short summary, and the TL;DR summary as long summary.
As a result, we make two sets of datasets: \textit{TIFU-short} and \textit{TIFU-long}.
Figure \ref{fig:post_example} shows an example post of the \texttt{TIFU} subreddit.

\begin{table}[t]
    \centering
    \small
    \setlength{\tabcolsep}{2.2pt}
    \begin{tabular}{|c|c|c|c|}
        \hline
        Dataset    & \# posts & \# words/post & \# words/summ \\
        \hline
        \texttt{RottenTomatoes} & 3,731 & 2124.7 (1747) & 22.2 (22) \\
        \texttt{Idebate}        & 2,259 & 178.3 (160)  & 11.4 (10) \\
        \hline
        {\tt TIFU-short} & 79,949   & 342.4 (269)  & 9.33 (8) \\
        {\tt TIFU-long } & 42,984   & 432.6 (351)  & 23.0 (21) \\
        \hline
    \end{tabular}
    \caption{
        Statistics of the Reddit TIFU dataset compared to existing opinion summarization corpora,
        \textit{RottenTomatoes} and \textit{Idebate} \cite{Wang:2016:NAACL-HLT}.
        We show average and median (in parentheses) values.
    }
    \vspace{-5pt}
    \label{tab:dataset}
\end{table}

\begin{figure*}[t] \begin{center}
    \includegraphics[width=\linewidth]{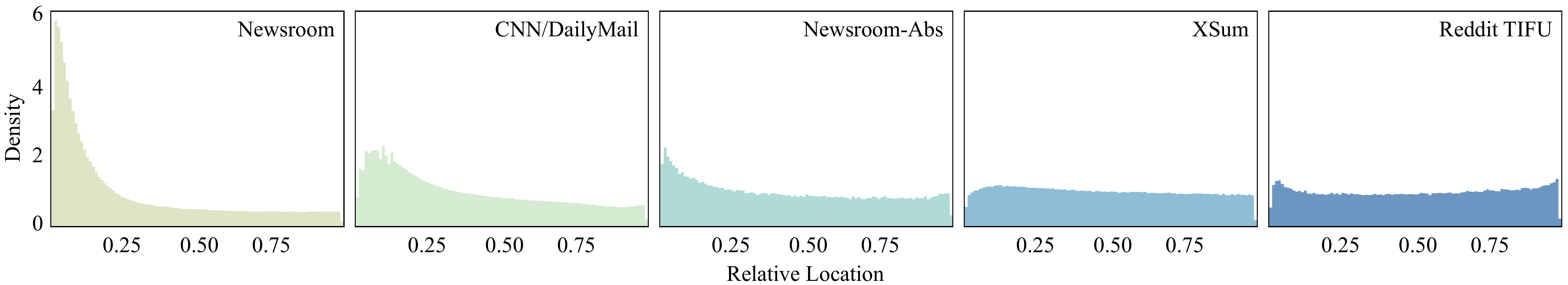}
    \caption{Relative locations of bigrams of gold summary in the source text across different datasets.}
    \label{fig:dataset_comparison}
    \vspace{-5pt}
\end{center} \end{figure*}

\begin{savenotes}
\begin{table*}[t!]
    \centering
    \small
    \setlength{\tabcolsep}{3.5pt}
    \begin{tabular}{|c|ccc|ccc|ccc|c|c|}
        \hline
                             & \multicolumn{3}{c|}{\texttt{PG}} 
                             & \multicolumn{3}{c|}{\texttt{Lead}}               & \multicolumn{3}{c|}{\texttt{Ext-Oracle}}        & \texttt{PG/Lead} & \texttt{PG/Oracle} \\
        \hline
        Dataset              & R-1             & R-2           & R-L             & R-1             & R-2           & R-L             & R-1            & R-2            & R-L            & Ratio (R-L)        & Ratio (R-L)    \\
        \hline
        {\tt CNN/DM}~\cite{Nallapati:2016:CoNLL}  & 36.4            & 15.7          & 33.4            & 39.6            & 17.7          & 36.2            & 54.7           & 30.4           & 50.8           & 0.92x              & 0.66x        \\ 
        {\tt NY Times}~\cite{Sandhaus:2008:LDC}       & 44.3            & 27.4          & 40.4            & 31.9            & 15.9          & 23.8            & 52.1           & 31.6           & 46.7           & 1.70x              & 0.87x       \\  
        {\tt Newsroom}~\cite{Grusky:2018:NAACL-HLT}       & 26.0            & 13.3          & 22.4            & 30.5            & 21.3          & 28.4            & 41.4           & 24.2           & 39.4           & 0.79x              & 0.57x        \\
        \hline
        {\tt Newsroom-Abs}~\cite{Grusky:2018:NAACL-HLT}   & \textbf{14.7}   & \textbf{2.2}  & \textbf{10.3}   & 13.7            & 2.4           & 11.2            & 29.7           & 10.5           & 27.2           & 0.92x              & 0.38x        \\
        {\tt XSum}~\cite{Narayan:2018:EMNLP}           & 29.7            & 9.2           & 23.2            & 16.3            & 1.6           & 12.0            & 29.8           & 8.8            & 22.7           & 1.93x              & 1.02x        \\
        \hline
        {\tt TIFU-short}     & 18.3            & 6.5           & 17.9            & \textbf{3.4}    & \textbf{0.0}  & \textbf{3.3}    & \textbf{8.0}   & \textbf{0.0}   & \textbf{7.7}   & \textbf{5.42x}     & \textbf{2.32x}  \\
        {\tt TIFU-long }     & 19.0            & 3.7           & 15.1            & \textbf{2.8}    & \textbf{0.0}  & \textbf{2.7}    & \textbf{6.8}   & \textbf{0.0}   & \textbf{6.6}   & \textbf{5.59x}     & \textbf{2.29x}  \\
        \hline
    \end{tabular}
    \caption{
        Comparison of F1 ROUGE scores between different datasets (row) and methods (column). 
\texttt{PG} is a state-of-the-art abstractive summarization method, and \texttt{Lead} and \texttt{Ext-Oracle} are extractive ones.
\texttt{PG/Lead} and \texttt{PG/Oracle} are the  ROUGE-L ratios of \texttt{PG} with \texttt{Lead} and \texttt{Ext-Oracle}, respectively.
We report the numbers for each dataset (row) from the corresponding cited papers.
    }
    \vspace{-5pt}
    \label{tab:dataset-ext}
\end{table*}
\end{savenotes}

\subsection{Preprocessing}
\label{sec:preprocessing}
We build a vocabulary dictionary $\mathcal{V}$ by choosing the most frequent $V$(=15K) words in the dataset.
\new{
We exclude any urls, unicodes and special characters.
We lowercase words, and normalize digits to 0.
Subreddit names and user ids are replaced with @subreddit and @userid token, respectively.
We use \texttt{markdown}\footnote{\url{https://python-markdown.github.io/}.} package to strip markdown format, and \texttt{spacy}\footnote{\url{https://spacy.io}.} to tokenize words.
Common prefixes of summary sentences (\eg tifu by, tifu-, tl;dr, etc) are trimmed.
We do not take OOV words into consideration, since our vocabulary with size 15K covers about 98\% of word frequencies in our dataset.
}
We set the maximum length of a document as 500. 
We exclude the gold summaries whose lengths are more than 20 and 50 for \textit{TIFU-short} and \textit{TIFU-long}, respectively. 
They amount to about 0.6K posts in both datasets (\ie less than 1\% and 3\%).
We use these maximum lengths, based on previous datasets (\eg 8, 31, 56 words on average per summary in
Gigaword, DUC, and CNN/DailyMail datasets, respectively).
We randomly split the dataset into 95\% for training, 5\% for test.

\subsection{Abstractive Properties of Reddit TIFU}
\label{sec:how_abstractive}

We discuss some abstractive characteristics found in Reddit TIFU dataset, compared to existing summarization datasets based on news articles.

\textbf{Weak Lead Bias}.
Formal documents including news articles tend to be structured to emphasize key information at the beginning of the text.
On the other hand, key information in informal online text data are more spread across the text.
Figure \ref{fig:dataset_comparison} plots the density histogram of the relative locations of bigrams of gold summary in the source text.
In the CNN/DailyMail and Newsroom, the bigrams are highly concentrated on the front parts of documents. 
Contrarily, our Reddit TIFU dataset shows rather uniform distribution across the text.

This characteristic can be also seen from the ROUGE score comparison in Table \ref{tab:dataset-ext}.
The \texttt{Lead} baseline  simply creates a summary by selecting the first few sentences or words in the document.
Thus, a high score of the Lead baseline implicates a strong lead bias.
The \texttt{Lead} scores are the lowest in our TIFU dataset, in which
it is more difficult for models to simply take advantage of locational bias for the summary.

\begin{figure*}[t] \begin{center}
    \includegraphics[width=\linewidth]{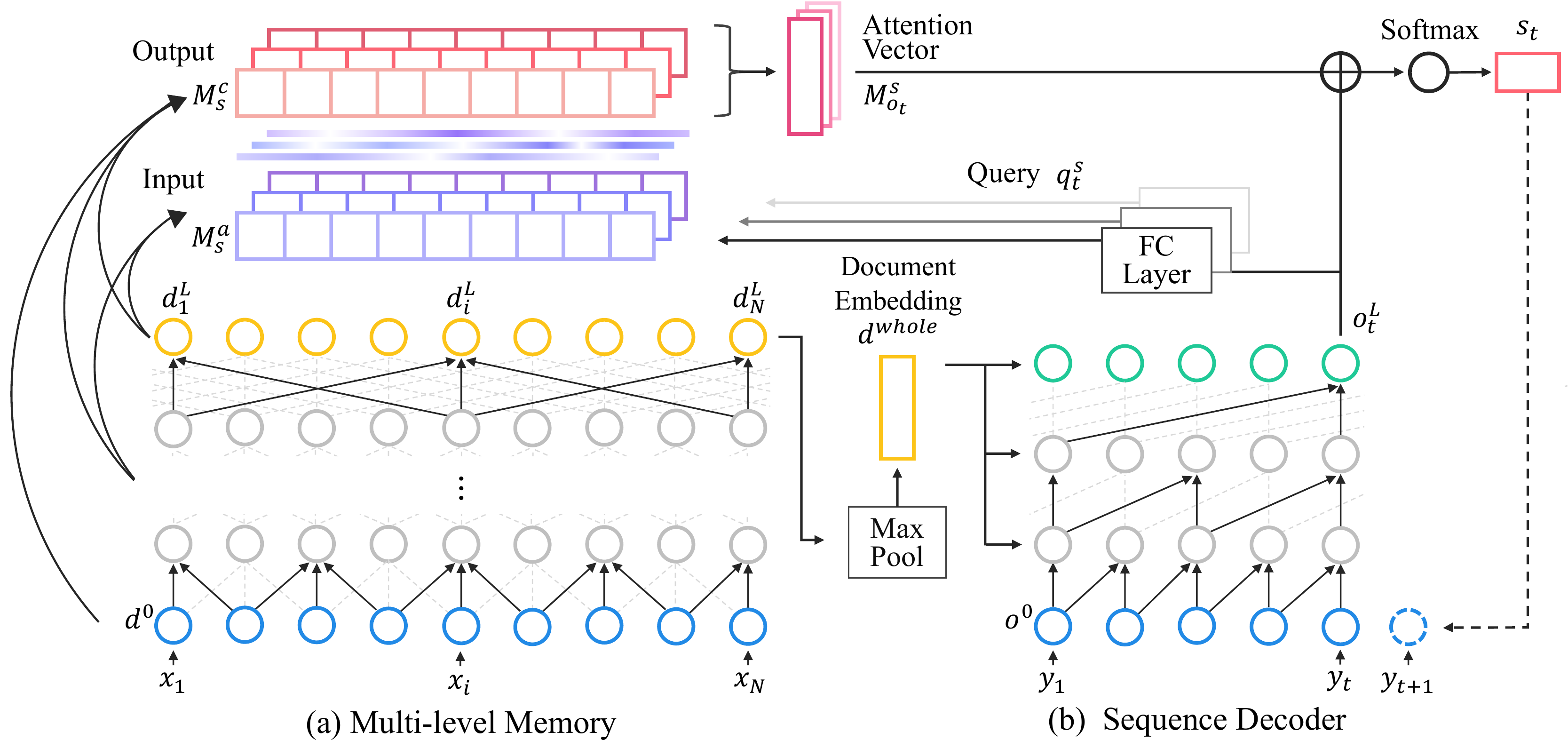}
    \caption{Illustration of the proposed \textit{multi-level memory network} (MMN) model.}
    \vspace{-5pt}
    \label{fig:model}
\end{center} \end{figure*}

\begin{figure}[t] \begin{center}
    \includegraphics[width=\linewidth]{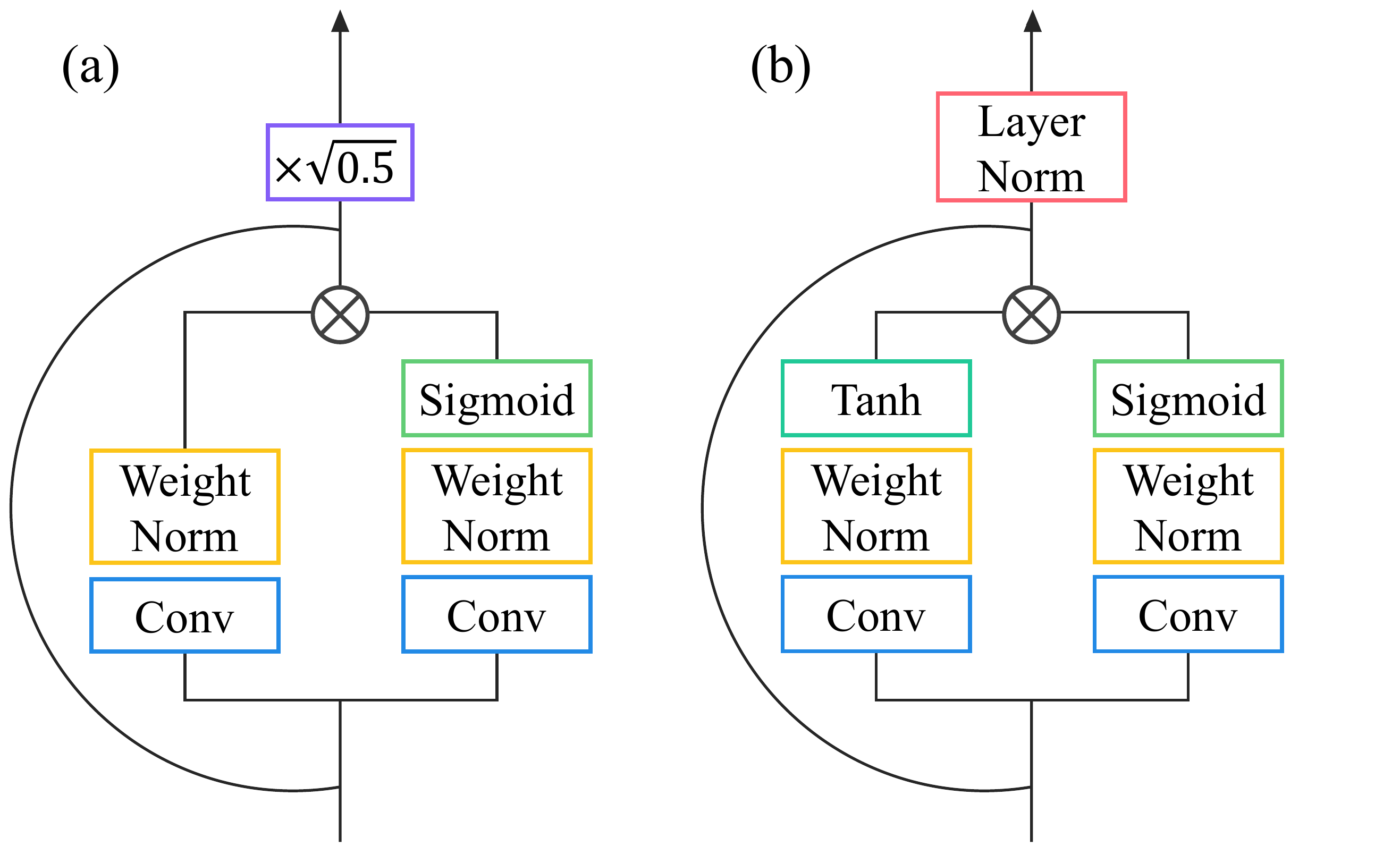}
    \caption{Comparison between (a) the gated linear unit \cite{Gehring:2017:ICML} and (b) the proposed normalized gated tanh unit.}
    \label{fig:gated_tanh}
\end{center} \vspace{-5pt} \end{figure}

\textbf{Strong Abstractness}.
Besides the locational bias, news articles tend to contain wrap-up sentences that cover the whole article,
and they often have resemblance to its gold summary.
Its existence can be measured by the score of the \texttt{Ext-Oracle} baseline, which
creates a summary by selecting the sentences with the highest average score of \new{F1} ROUGE-1/2/L.
Thus, it can be viewed as an upper bound for extractive models \cite{Narayan:2018:EMNLP, Narayan:2018:NAACL-HLT, Nallapati:2017:AAAI}.

In Table \ref{tab:dataset-ext}, the ROUGE scores of the \texttt{Ext-Oracle} are the lowest in our TIFU dataset.
It means that  the sentences that are similar to gold summary scarcely exist inside the source text  in our dataset.
This property forces the model to be trained to focus on comprehending the entire text instead of simply finding wrap-up sentences.

Finally, \texttt{PG/Lead} and \texttt{PG/Oracle} in Table \ref{tab:dataset-ext} are the  ROUGE-L ratios of \texttt{PG} with \texttt{Lead} and \texttt{Ext-Oracle}, respectively.
These metrics can quantify the dataset according to the degree of difficulty for extractive methods and the suitability for abstractive methods, respectively. 
High scores of the TIFU dataset in both metrics show that it is potentially an excellent benchmark for evaluation of abstractive summarization systems. 

\section{Multi-level Memory Networks (MMN)}
\label{sec:model}

Figure \ref{fig:model} shows the proposed \textit{multi-level memory network} (MMN) model. 
The MMN memorizes the source text with a proper representation in the memory
and generates a summary sentence one word at a time by extracting relevant information from memory cells in response to previously generated words. 
The input of the model is a source text $\{x_i\} = x_1, ..., x_N$, and the output is a sequence of summary words  $\{y_t\} = y_1, ..., y_T$, each of which is a symbol from the dictionary $\mathcal V$.

\subsection{Text Embedding}
\label{sec:embedding_layers}

Online posts include lots of morphologically similar words, which should be closely embedded.
Thus, we use the \texttt{fastText} \cite{Bojanowski:2016:TACL} trained on the Common Crawl corpus, 
to initialize the word embedding matrix $\mathbf W_{emb}$.
We use the same embedding matrix $\mathbf W_{emb}$ for both source text and output sentences.
That is, we represent a source text $\{x_i\}_{i=1}^N$ in a distributional space as $\{\mathbf d_i^0\}_{i=1}^N $ by 
$\mathbf d_i^0 = \mathbf W_{emb} \mathbf x_i$  where $\mathbf x_i$ is a one-hot vector for $i$-th word in the source text. 
Likewise,  output words $\{y_t\}_{t=1}^T$ is embedded as $\{\mathbf o_t^0\}_{t=1}^T$,  
and $\mathbf d_i^0$ and $\mathbf o_t^0 \in \mathbb R^{300}$.

\subsection{Construction of Multi-level Memory}
\label{sec:construct_memory}

As shown in Figure \ref{fig:model}(a), 
the multi-level memory network takes the source text embedding $\{\mathbf d_i^0\}_{i=1}^N$ as an input,
and generates $S$ number of memory tensors $\{\mathbf M_s^{a/c}\}_{s=1}^S$ as output,
where superscript $a$ and $c$ denote input and output memory representation, respectively.
The multi-level memory network is motivated by that when human understand a document, she does not remember it as a single whole document
but ties together several levels of abstraction (\eg word-level, sentence-level, paragraph-level and document-level).
That is, we generate $S$ sets of memory tensors, each of which associates each cell with different number of neighboring word embeddings based on the level of abstraction. 
To build memory slots of such multi-level memory,  we exploit a multi-layer CNN as the write network, where each layer is chosen based on the size of its receptive field.

However, one issue of convolution is that large receptive fields require many layers or large filter sizes.
For example, stacking 6 layers with a filter size of 3 results in a receptive field size of 13, \ie each output depends on 13 input words.
In order to grow the receptive field without increasing the computational cost, we exploit the \textit{dilated} convolution~\cite{Yu:2016:ICLR,Oord:2016:SSW} for the write network.

\textbf{Memory Writing with Dilated Convolution}.
In dilated convolution, the filter is applied over an area larger than its length by skipping input values with a certain gap.
Formally, for a 1-D $n$-length input $\mathbf x \in \mathbb{R}^{n \times 300}$ and a filter $\mathbf w : \{1, ..., k\} \rightarrow \mathbb{R}^{300}$, the dilated convolution operation $\mathcal{F}$ on $s$ elements of a sequence is defined as
    \begin{align}
        \label{eq:dilated_cnn_op}
        \mathcal{F} (\mathbf x, s) = \sum_{i=1}^{k} \mathbf w (i) * \mathbf x_{s + d \cdot (i - \lfloor k/2 \rfloor)} + \mathbf b, 
    \end{align}
where $d$ is the dilation rate, $k$ is the filter size, $s - d \cdot (i - \lfloor k/2 \rfloor)$ accounts for the direction of dilation and
$\mathbf w \in \mathbb R^{k \times 300 \times 300}$ and $\mathbf b \in \mathbb R^{300}$ are the parameters of the filter.
With $d = 1$, the dilated convolution reduces to a regular convolution.
Using a larger dilation enables a single output at the top level to represent a wider range of input, thus effectively expanding the receptive field.

To the embedding of a source text $\{\mathbf d_i^0\}_{i=1}^N$, we recursively apply a series of dilated convolutions 
$F(\mathbf d^0) \in \mathbb R^{N \times 300}$.
We denote the output of the $l$-th convolution layer as $\{\mathbf d_i^l\}_{i=1}^N$. 

\textbf{Normalized Gated Tanh Units}.
Each convolution is followed by our new activation of \textit{normalized gated tanh unit} (NGTU), which is illustrated in Figure \ref{fig:gated_tanh}(b):
    \begin{align}
        \label{eq:gtu}
        \mbox{GTU}(\mathbf d^l) &= \mbox{tanh} (\mathcal{F}_f^l (\mathbf d^l)) \circ \sigma (\mathcal{F}_g^l (\mathbf d^l)), \\
        \label{eq:residual}
        \mathbf d^{l+1} &= \mbox{LayerNorm} (\mathbf d^l + \mbox{GTU}(\mathbf d^l)),
    \end{align}
where $\sigma$ is a sigmoid, $\circ$ is the element-wise multiplication and $F_f^l$ and $F_g^l$ denote the filter and gate for $l$-th layer dilated convolution, respectively.

The NGTU is an extension of the existing gated tanh units (GTU) \cite{Oord:2016:SSW, Oord:2016:NIPS} by applying weight normalization \cite{Salimans:2016:NIPS} and layer normalization \cite{Ba:2016:Stat}.
This mixed normalization improves earlier work of \citet{Gehring:2017:ICML}, where only weight normalization is applied to the GLU.
As in Figure \ref{fig:gated_tanh}(a), it tries to preserve the variance of activations throughout the whole network by scaling the output of residual blocks by $\sqrt{0.5}$.
However, we  observe that this heuristic does not always preserve the variance and does not empirically work well in our dataset.
Contrarily, the proposed NGTU not only guarantees preservation of activation variances but also  significantly improves the performance.

\textbf{Multi-level Memory}.
Instead of using only the last layer output of CNNs, we exploit the outputs of multiple layers of CNNs to construct $S$ sets of memories.
For example, memory constructed from the 4-th layer, whose receptive field is 31, may have sentence-level embeddings, while memory from the 8-th layer, whose receptive field is 511, may have document-level embeddings.
We obtain each $s$-th level memory $\mathbf M_s^{a/c}$ by resembling key-value memory networks \cite{Miller:2016:EMNLP}:
\begin{align}
   \mathbf M_s^a &= \mathbf d^{\mathbf m(s)}, ~ \mathbf M_s^c = \mathbf d^{\mathbf m(s)} + \mathbf d^0.
\end{align}
Recall that $\mathbf M_s^a$ and $\mathbf M_s^c \in \mathbb R^{N \times 300}$ are input and output memory matrix, respectively.
$\mathbf {m} (s)$ indicates an index of convolutional layer used for the $s$-th level memory.
For example, if we set $S=3$ and $\mathbf m = \{3,6,9 \}$, we make three-level memories, each of which uses the output of the 3-rd, 6-th, and 9-th convolution layer, respectively. 
To output memory representation $\mathbf M_s^c$, we add the document embedding $\mathbf d^{0}$ as a skip connection. 

\subsection{State-Based Sequence Generation}
\label{sec:query_network}

We discuss how to predict the next word $y_{t+1}$ at time step $t$ based on the memory state and previously generated words $y_{1:t}$. 
Figure \ref{fig:model}(b) visualizes the overall procedure of decoding. 

We first apply max-pooling to the output of the last layer of the encoder network to build a whole document embedding $\mathbf d^{whole} \in \mathbb R^{300}$:
\begin{align}
    \label{eq:document_max_pooling}
    \mathbf d^{whole} = \mbox{maxpool}([\mathbf d^L_1; ...; \mathbf d^L_N]).
\end{align}

The decoder is designed based on WaveNet \cite{Oord:2016:SSW} that uses a series of causal dilated convolutions, denoted by $\hat{\mathcal{F}}(\mathbf o_{1:t}^l) \in \mathbb R^{t \times 300}$.
We globally condition $\mathbf d^{whole}$
to obtain embeddings of previously generated words $\mathbf o_{1:t}^l$ as: 
\begin{align}
    &\mathbf h_{f/g}^l = \hat{\mathcal{F}}_{f/g}^l ( \mathbf o^l_{1:t} + \mathbf W_{f/g}^l \mathbf d^{whole}), \\
    &\mathbf h_a^l 
                  = \mbox{tanh} (\mathbf h_f^l) \circ \sigma (\mathbf h_g^l), \\
    \label{eq:output_conv}
    &\mathbf o^{l+1}_{1:t} = \mbox{LayerNorm} (\mathbf o^l_{1:t} + \mathbf h_a^l),
\end{align}
where $\mathbf h_{f/g}^l$ are the filter and gate hidden state respectively, and learnable parameters are $\mathbf W_f^l$ and $\mathbf W_g^l \in \mathbb R^{300 \times 300}$.
We initialize $\mathbf o^0_t = \mathbf W_{emb} \mathbf y_t$.
We set the level of the decoder network to $L = 3$ for TIFU-short and $L = 5$ for TIFU-long.

Next, we generate $S$ number of query vectors $\{\mathbf q_t^s\}_{s=1}^S$ at time $t$ to our memory network as
\begin{align}
    \label{eq:query}
    \mathbf q_t^s &= \mbox{tanh}(\mathbf W_q^s \mathbf o^L_t + \mathbf b_q^s),
\end{align}
where $\mathbf W_q^s \in \mathbb R^{300 \times 300}$ and $\mathbf b_q^s \in \mathbb R^{300}$.

Each of these query vectors $\{\mathbf q_t^s\}_{s=1}^S$ is fed into the attention function of each level of memory.
As in \citep{Vaswani:2017:NIPS}, the attention function is
\begin{align}
    \label{eq:attention}
    \mathbf M_{o_t}^s = \mbox{softmax}(\frac{\mathbf q_t^s (\mathbf M_s^a)^T}{\sqrt{d^{emb}}}) \mathbf M_s^c,
\end{align}
where we set $d^{emb}=300$ for the embedding dimension and $\mathbf M_{o_t}^s \in \mathbb R^{300}$.

Next, we obtain the output word probability: 
\begin{align}
    \label{eq:output_act}
    \mathbf s_t = \mbox{softmax}(\mathbf W_o [\mathbf M_{o_t}^1;...;\mathbf M_{o_t}^S;\mathbf o^L_t]),
\end{align}
where $\mathbf W_o \in \mathbb R^{(300 \times (S + 1)) \times V}$.
Finally, we select the word with the highest probability $y_{t+1} = \mbox{argmax}_{\mathbf s \in \mathcal{V}} (\mathbf s_t)$.
Unless $y_{t+1}$ is an EOS token, we repeat generating the next word by feeding $y_{t+1}$ into the output convolution layer of Eq.(\ref{eq:output_conv}).

\subsection{Training}
\label{sec:training}

We use the softmax cross-entropy loss from estimated $y_t$ to its target $y_{GT,t}$. 
However, it forces the model to predict extremes (zero or one) to distinguish among the ground truth and alternatives.
The label smoothing alleviates this issue by acting as a regularizer that makes the model less confident in its prediction.
We smooth the target distribution with a uniform prior distribution $u$ \cite{Pereyra:2017:ICLR, Edunov:2017:NAACL-HLT, Vaswani:2017:NIPS}.
Thus, the loss over the training set $\mathcal{D}$ is
    \begin{align}
       \mathcal{L} = - \sum \log p_{\theta} (\mathbf{y} | \mathbf{x}) - D_{KL}(u||p_{\theta}(\mathbf{y} | \mathbf{x})). \nonumber
    \end{align}
We implement label smoothing by modifying the ground truth distribution for word $y_{GT,t}$ to be $p(y_{GT,t}) = 1 - \epsilon$
and $p(y') = \epsilon/ \mathcal{V}$ for $y' \neq y_{GT,t}$ where $\epsilon$ is a smoothing parameter set to 0.1.
Further details can be found in the Appendix. 

\section{Experiments}
\label{sec:experiments}

\subsection{Experimental Setting}
\label{sec:experimental_setting}

\textbf{Evaluation Metrics}.
We evaluate the summarization performance with two language metrics: perplexity and standard F1 ROUGE scores \cite{Lin:2004:TSBO}. 
We remind that lower perplexity and higher ROUGE scores indicate better performance. 

\textbf{Datasets}.
In addition to Reddit TIFU, we also evaluate on two existing datasets: abstractive subset of \texttt{Newsroom} \cite{Grusky:2018:NAACL-HLT} and \texttt{XSum} \cite{Narayan:2018:EMNLP}.
These are suitable benchmarks for evaluation of our model in two aspects.
First, they are specialized for abstractive summarization, which meets well the goal of this work. 
Second, they have larger vocabulary size (40K, 50K) than Reddit TIFU (15K), and thus we can evaluate the learning capability of our model.

\textbf{Baselines}.
We compare with three abstractive summarization methods, one basic seq2seq model,
two heuristic extractive methods and variants of our model.
We choose \texttt{PG} \cite{See:2017:ACL}, \texttt{SEASS} \cite{Zhou:2017:ACL}, \texttt{DRGD} \cite{Li:2017:EMNLP} as the state-of-the-art methods of abstractive summarization.
We test the attention based seq2seq model denoted as \texttt{s2s-att} \cite{Chopra:2016:NAACL-HLT}.
As heuristic extractive methods, the \texttt{Lead-1} uses the first sentence in the text  as summary, and the \texttt{Ext-Oracle} takes the sentence with the highest average score of F1 ROUGE-1/2/L with the gold summary in the text.
Thus, \texttt{Ext-Oracle} can be viewed as an upper-bound for extractive methods.

We also test variants of our method \texttt{MMN-*}. To validate the contribution of each component, we exclude one of key components from our model as follows: 
(i) \texttt{-NoDilated} with conventional convolutions instead,
(ii) \texttt{-NoMulti}  with no multi-level memory
(iii) \texttt{-NoNGTU} with existing gated linear units \cite{Gehring:2017:ICML}.
That is, \texttt{-NoDilated} quantifies the improvement by the dilated convolution,
\texttt{-NoMulti} assesses the effect of multi-level memory,
and \texttt{-NoNGTU} validates the normalized gated tanh unit.

Please refer to the Appendix for implementation details of our method.

\begin{table}[t]
    \centering
    \small
    \setlength{\tabcolsep}{4pt}
    \begin{tabular}{|c|c|ccc|}
        \hline
        \multicolumn{5}{|c|}{\textbf{TIFU-short}} \\
        \hline
        Methods                                    & PPL       & R-1     & R-2     & R-L      \\ 
        \hline
        {\tt Lead-1}                               & n/a       & 3.4     & 0.0     & 3.3      \\
        {\tt Ext-Oracle}                           & n/a       & 8.0     & 0.0     & 7.7      \\ 
        \hline
        {\tt s2s-att} \cite{Chopra:2016:NAACL-HLT}                          & 46.2      & 18.3    & 6.4     & 17.8     \\ 
        {\tt PG}  \cite{See:2017:ACL}              & 40.9      & 18.3    & 6.5     & 17.9     \\ 
        {\tt SEASS} \cite{Zhou:2017:ACL}           & 62.6      & 18.5    & 6.4     & 18.0     \\ 
        {\tt DRGD} \cite{Li:2017:EMNLP}            & 69.2      & 14.6    & 3.3     & 14.2     \\ 
        \hline
        {\tt MMN}                                  & 32.1 & \textbf{20.2} & \textbf{7.4} & \textbf{19.8} \\ 
        {\tt MMN-NoDilated}                        & \textbf{31.8}  & 19.5    & 6.8     & 19.1  \\
        {\tt MMN-NoMulti}                          & 34.4      & 19.0    & 6.1     & 18.5  \\
        {\tt MMN-NoNGTU}                           & 40.8      & 18.6    & 5.6     & 18.1  \\
        \hline
        \multicolumn{5}{|c|}{\textbf{TIFU-long}} \\
        \hline
        {\tt Lead-1}                               & n/a       & 2.8     & 0.0     & 2.7      \\
        {\tt Ext-Oracle}                           & n/a       & 6.8     & 0.0     & 6.6      \\ 
        \hline
        {\tt s2s-att} \cite{Chopra:2016:NAACL-HLT} & 180.6     & 17.3    & 3.1     & 14.0     \\ 
        {\tt PG} \cite{See:2017:ACL}               & 175.3     & 16.4    & 3.0     & 13.5     \\ 
        {\tt SEASS} \cite{Zhou:2017:ACL}           & 387.0     & 17.5    & 2.9     & 13.9     \\ 
        {\tt DRGD} \cite{Li:2017:EMNLP}            & 176.6     & 16.8    & 2.0     & 13.6     \\ 
        \hline
        {\tt MMN}                                  & \textbf{114.1} & \textbf{19.0} & \textbf{3.7} & \textbf{15.1} \\ 
        {\tt MMN-NoDilated}                        & 124.2 & 17.6   & 3.4   & 14.1  \\
        {\tt MMN-NoMulti}                          & 124.5      & 14.0   & 1.5   & 11.8  \\
        {\tt MMN-NoNGTU}                           & 235.4      & 14.0   & 2.6   & 12.1  \\
        \hline
    \end{tabular}
    \caption{
        Summarization results measured by perplexity and ROUGE-1/2/L on the TIFU-short/long dataset.
    }
    \label{tab:results}
\end{table}

\begin{table}[t]
    \centering
    \small
    \setlength{\tabcolsep}{4pt}
    \begin{tabular}{|c|ccc|ccc|}
        \hline
        & \multicolumn{3}{c|}{\textbf{Newsroom-Abs} } & \multicolumn{3}{c|}{\textbf{XSum} } \\
        \hline
        Methods             & R-1           & R-2           & R-L           & R-1           & R-2           & R-L           \\
        \hline
        {\tt s2s-att}       & 6.2           & 1.1           & 5.7           & 28.4          & 8.8           & 22.5          \\
        {\tt PG}            & 14.7          & 2.2           & 11.4          & 29.7          & 9.2           & 23.2          \\
        {\tt ConvS2S}       & -             & -             & -             & 31.3          & 11.1          & 25.2          \\
        {\tt T-ConvS2S}     & -             & -             & -             & 31.9          & 11.5          & 25.8          \\
        {\tt MMN} (Ours)    & \textbf{17.5} & \textbf{4.7}  & \textbf{14.2} & \textbf{32.0} & \textbf{12.1} & \textbf{26.0} \\
        \hline
    \end{tabular}
    \caption{
        Summarization results in terms of ROUGE-1/2/L on Newsroom-Abs \cite{Grusky:2018:NAACL-HLT} and XSum \cite{Narayan:2018:EMNLP}.
        Except \texttt{MMN}, all scores are referred to the original papers. {\tt T-ConvS2S}  is the topic-aware convolutional seq2seq model.
    }
    \vspace{-5pt}
    \label{tab:results_newsroom}
\end{table}

\subsection{Quantitative Results}
\label{sec:results_word_overlap}

Table \ref{tab:results} compares the summarization performance of different methods on the TIFU-short/long dataset. 
Our model outperforms the state-of-the-art abstractive methods in both ROUGE and perplexity scores.
\texttt{PG} utilizes a pointer network to copy words from the source text, but it may not be a good strategy in our dataset,
which is more abstractive as discussed in Table \ref{tab:dataset-ext}. 
\texttt{SEASS} shows strong performance in DUC and Gigaword dataset, 
in which the source text is a single long sentence and the gold summary is its shorter version. 
Yet, it may not be sufficient to summarize much longer articles of our dataset, even with its second-level representation.
\texttt{DRGD} is based on the variational autoencoder with latent variables to capture the structural patterns of gold summaries. 
This idea can be useful for the similarly structured formal documents 
but may not go well with diverse online text in the TIFU dataset.

\begin{table}[t]
    \small
    \centering
    \setlength{\tabcolsep}{4.5pt}
\begin{tabular}{|c|cc|c|cc|c|}
        \hline
        & \multicolumn{3}{c|}{\textbf{TIFU-short}} & \multicolumn{3}{c|}{\textbf{TIFU-long}}\\
        \hline
        vs. Baselines & Win & Lose & Tie & Win & Lose & Tie  \\
        \hline
        \texttt{s2s-att} & \textbf{43.0} & 28.3 & 28.7 & \textbf{32.0} & 24.0 & 44.0 \\
        \texttt{PG}          & \textbf{38.7} & 28.0 & 33.3 & \textbf{42.3} & 33.3 & 24.3 \\
        \texttt{SEASS}       & \textbf{35.7} & 28.0 & 36.3 & \textbf{47.0} & 37.3 & 15.7 \\
        \texttt{DRGD}        & \textbf{46.7} & 17.3 & 15.0 & \textbf{61.0} & 23.0 & 16.0 \\
        \hline
        Gold    & 27.0 & \textbf{58.0} & 15.0 & 22.3 & \textbf{73.7} & 4.0 \\
        \hline
    \end{tabular}
    \caption{
        AMT results on the TIFU-short/long between our \texttt{MMN} and four baselines and gold summary.
        We show percentages of responses that turkers vote for our approach over baselines.
    }
    \vspace{-5pt}
    \label{tab:results_amt}
\end{table}

These state-of-the-art abstractive methods are not as good as our model, but still perform better than extractive methods.
Although the \texttt{Ext-Oracle} heuristic is an upper-bound for extractive methods, 
it is not successful in our highly abstractive dataset; it is not effective to simply retrieve existing sentences from the source text.
Moreover, the performance gaps between abstractive and extractive methods are much larger in our dataset than in other datasets \cite{See:2017:ACL, Paulus:2018:ICLR, Cohan:2018:NAACL-HLT}, which means too that our dataset is highly abstractive.

Table \ref{tab:results_newsroom} compares the performance of our MMN on Newsroom-Abs and XSum dataset.
We report the numbers from the original papers.
Our model outperforms not only the RNN-based abstractive methods but also the convolutional-based methods in all ROUGE scores.
Especially, even trained on single end-to-end training procedure, our model outperforms \texttt{T-ConvS2S}, which necessitates two training stages of LDA and \texttt{ConvS2S}.
These results assure that even on formal documents with large vocabulary sizes, our multi-level memory is effective for abstractive datasets.

\subsection{Qualitative Results}
\label{sec:qualitative_results}

\begin{figure}[t] \begin{center}
    \includegraphics[width=\linewidth]{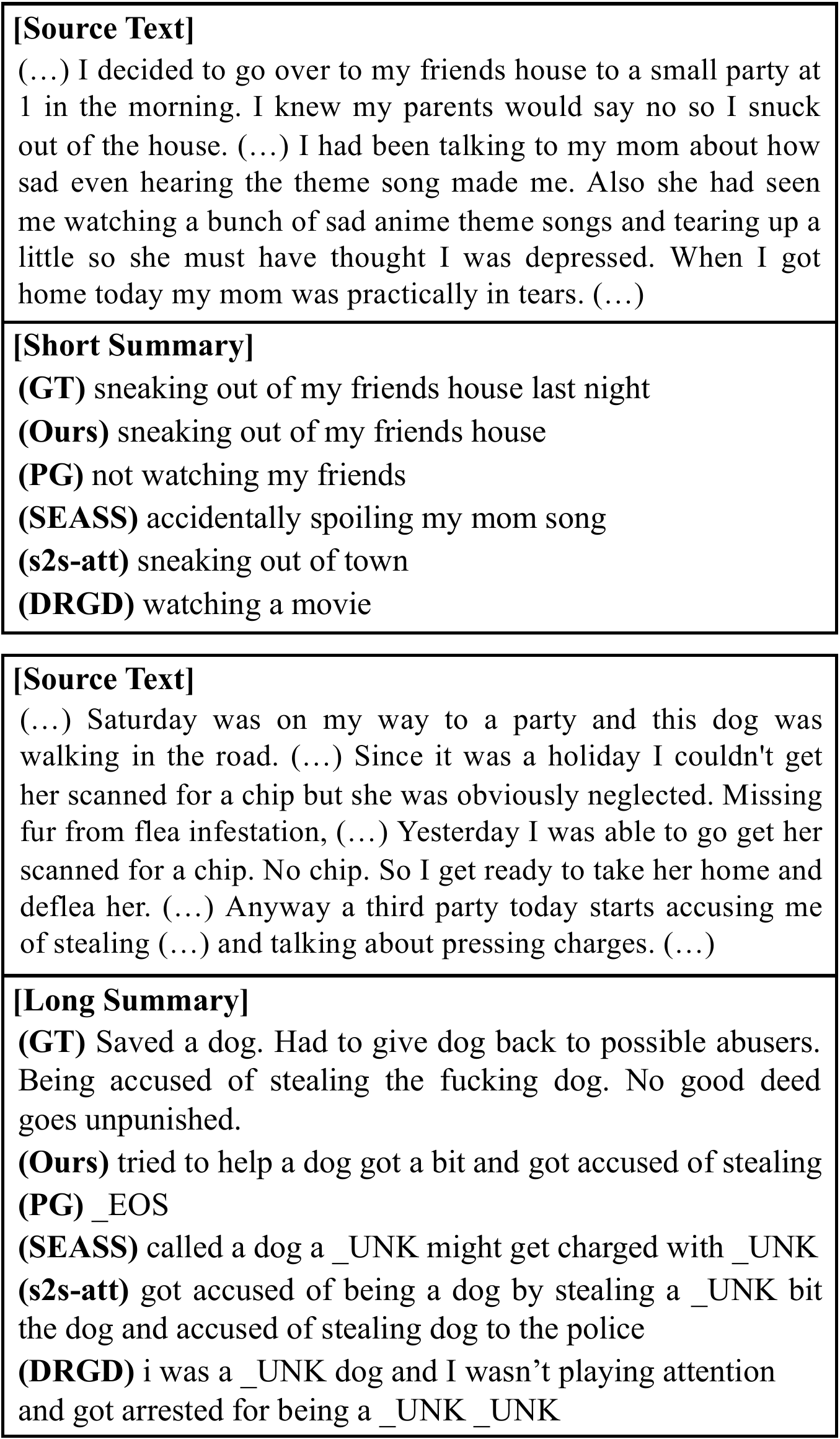}
    \caption{Examples of abstractive summary generated by our model and baselines.
    In each set, we too show the source text and gold summary.} 
    \label{fig:example_short}
    \vspace{-5pt}
\end{center} \end{figure}

We perform two types of qualitative evaluation to complement the limitation of automatic language metrics as summarization evaluation. 

\textbf{User Preferences}.
We perform Amazon Mechanical Turk (AMT) tests to observe general users' preferences between the summarization of different algorithms. 
We randomly sample 100 test examples.
At test, we show a source text and two summaries generated by our method and one baseline in a random order.
We ask turkers to choose the more relevant one for the source text.
We obtain answers from three different turkers for each test example.
We compare with four abstractive baselines (\texttt{s2s-att}, \texttt{PG}, \texttt{SEASS} and \texttt{DRGD}) and the gold summary (Gold).

Table \ref{tab:results_amt} summarizes the results of AMT tests, which validate that human annotators significantly prefer our results to those of baselines.
As expected, the gold summary is voted the most.

\textbf{Summary Examples}.
Figure \ref{fig:example_short} shows selected examples of abstractive summarization.
Baselines often generate the summary by mostly focusing on some keywords in the text, while our model produces the summary considering both keywords and the whole context thanks to multi-level memory.
We present more examples in the Appendix.

\section{Conclusions}
\label{sec:conclusion}

We introduced a new dataset \textit{Reddit TIFU} for abstractive summarization on informal online text.
We also proposed a novel summarization model named \textit{multi-level memory networks} (MMN). 
Experiments showed that the Reddit TIFU dataset is uniquely abstractive and the MMN model is highly effective. 
There are several promising future directions. 
First, ROUGE metrics are limited to correctly capture paraphrased summaries, for which a new automatic metric of abstractive summarization may be required.
Second, we can explore the data in other online forums such as Quora, Stackoverflow and other subreddits.

\section*{Acknowledgments}
\label{sec:acknowledgments}

We thank Chris Dongjoo Kim, Yunseok Jang and the anonymous reviewers for their helpful comments.
This work was supported by Kakao and Kakao Brain corporations and IITP grant funded by the Korea government (MSIT) (No. 2017-0-01772, Development of QA systems for Video Story Understanding to pass the Video Turing Test). 
Gunhee Kim is the corresponding author.

\bibliography{naaclhlt2019_summarization}

\begin{thebibliography}{50}
\expandafter\ifx\csname natexlab\endcsname\relax\def\natexlab#1{#1}\fi

\bibitem[{Allahyari et~al.(2017)Allahyari, Pouriyeh, Assefi, Safaei, Trippe,
  Gutierrez, and Kochut}]{Allahyari:2017:arXiv}
Mehdi Allahyari, Seyedamin Pouriyeh, Mehdi Assefi, Saeid Safaei, Elizabeth~D
  Trippe, Juan~B Gutierrez, and Krys Kochut. 2017.
\newblock {Text Summarization Techniques: A Brief Survey}.
\newblock In \emph{arXiv:1707.02268}.

\bibitem[{Ba et~al.(2016)Ba, Kiros, and Hinton}]{Ba:2016:Stat}
Jimmy~Lei Ba, Jamie~Ryan Kiros, and Geoffrey~E Hinton. 2016.
\newblock {Layer Normalization}.
\newblock In \emph{Stat}.

\bibitem[{Bojanowski et~al.(2016)Bojanowski, Grave, Joulin, and
  Mikolov}]{Bojanowski:2016:TACL}
Piotr Bojanowski, Edouard Grave, Armand Joulin, and Tomas Mikolov. 2016.
\newblock {Enriching Word Vectors with Subword Information}.
\newblock In \emph{TACL}.

\bibitem[{Celikyilmaz et~al.(2018)Celikyilmaz, Bosselut, He, and
  Choi}]{Celikyilmaz:2018:NAACL-HLT}
Asli Celikyilmaz, Antoine Bosselut, Xiaodong He, and Yejin Choi. 2018.
\newblock {Deep Communicating Agents for Abstractive Summarization}.
\newblock In \emph{NAACL-HLT}.

\bibitem[{Chopra et~al.(2016)Chopra, Auli, and Rush}]{Chopra:2016:NAACL-HLT}
Sumit Chopra, Michael Auli, and Alexander~M Rush. 2016.
\newblock {Abstractive Sentence Summarization With Attentive Recurrent Neural
  Networks}.
\newblock In \emph{NAACL-HLT}.

\bibitem[{Cohan et~al.(2018)Cohan, Dernoncourt, Kim, Bui, Kim, Chang, and
  Goharian}]{Cohan:2018:NAACL-HLT}
Arman Cohan, Franck Dernoncourt, Doo~Soon Kim, Trung Bui, Seokhwan Kim, Walter
  Chang, and Nazli Goharian. 2018.
\newblock {A Discourse-Aware Attention Model for Abstractive Summarization of
  Long Documents}.
\newblock In \emph{NAACL-HLT}.

\bibitem[{Edunov et~al.(2017)Edunov, Ott, Auli, Grangier, and
  Ranzato}]{Edunov:2017:NAACL-HLT}
Sergey Edunov, Myle Ott, Michael Auli, David Grangier, and Marc'Aurelio
  Ranzato. 2017.
\newblock {Classical Structured Prediction Losses for Sequence to Sequence
  Learning}.
\newblock In \emph{NAACL-HLT}.

\bibitem[{Gehring et~al.(2017)Gehring, Auli, Grangier, Yarats, and
  Dauphin}]{Gehring:2017:ICML}
Jonas Gehring, Michael Auli, David Grangier, Denis Yarats, and Yann~N Dauphin.
  2017.
\newblock {Convolutional Sequence to Sequence Learning}.
\newblock In \emph{ICML}.

\bibitem[{Gehrmann et~al.(2018)Gehrmann, Deng, and Rush}]{Gehrmann:2018:EMNLP}
Sebastian Gehrmann, Yuntian Deng, and Alexander Rush. 2018.
\newblock Bottom-up abstractive summarization.
\newblock In \emph{EMNLP}.

\bibitem[{Glorot and Bengio(2010)}]{Glorot:2010:AISTATS}
Xavier Glorot and Yoshua Bengio. 2010.
\newblock {Understanding the Difficulty of Training Deep Feedforward Neural
  Networks}.
\newblock In \emph{AISTATS}.

\bibitem[{Grusky et~al.(2018)Grusky, Naaman, and Artzi}]{Grusky:2018:NAACL-HLT}
Max Grusky, Mor Naaman, and Yoav Artzi. 2018.
\newblock {Newsroom: A Dataset of 1.3 Million Summaries with Diverse Extractive
  Strategies}.
\newblock In \emph{NAACL-HLT}.

\bibitem[{Hermann et~al.(2015)Hermann, Kocisky, Grefenstette, Espeholt, Kay,
  Suleyman, and Blunsom}]{Hermann:2015:NIPS}
Karl~Moritz Hermann, Tomas Kocisky, Edward Grefenstette, Lasse Espeholt, Will
  Kay, Mustafa Suleyman, and Phil Blunsom. 2015.
\newblock {Teaching Machines to Read and Comprehend}.
\newblock In \emph{NIPS}.

\bibitem[{Hsu et~al.(2018)Hsu, Lin, Lee, Min, Tang, and Sun}]{Hsu:2018:ACL}
Wan-Ting Hsu, Chieh-Kai Lin, Ming-Ying Lee, Kerui Min, Jing Tang, and Min Sun.
  2018.
\newblock {A Unified Model for Extractive and Abstractive Summarization using
  Inconsistency Loss}.
\newblock In \emph{ACL}.

\bibitem[{Hu et~al.(2015)Hu, Chen, and Zhu}]{Hu:2015:EMNLP}
Baotian Hu, Qingcai Chen, and Fangze Zhu. 2015.
\newblock Lcsts: A large scale chinese short text summarization dataset.
\newblock In \emph{EMNLP}.

\bibitem[{Kaiser et~al.(2017)Kaiser, Nachum, Roy, and
  Bengio}]{Kaiser:2017:ICLR}
{\L}ukasz Kaiser, Ofir Nachum, Aurko Roy, and Samy Bengio. 2017.
\newblock {Learning to Remember Rare Events}.
\newblock In \emph{ICLR}.

\bibitem[{Kedzie et~al.(2018)Kedzie, McKeown, and
  Daume~III}]{Kedzie:2018:EMNLP}
Chris Kedzie, Kathleen McKeown, and Hal Daume~III. 2018.
\newblock {Content Selection in Deep Learning Models of Summarization}.
\newblock In \emph{EMNLP}.

\bibitem[{Kingma and Ba(2015)}]{Kingma:2015:ICLR}
Diederik Kingma and Jimmy Ba. 2015.
\newblock {Adam: A Method for Stochastic Optimization}.
\newblock In \emph{ICLR}.

\bibitem[{Kumar et~al.(2016)Kumar, Irsoy, Su, Bradbury, English, Pierce,
  Ondruska, Gulrajani, and Socher}]{Kumar:2016:ICML}
Ankit Kumar, Ozan Irsoy, Jonathan Su, James Bradbury, Robert English, Brian
  Pierce, Peter Ondruska, Ishaan Gulrajani, and Richard Socher. 2016.
\newblock {Ask me Anything: Dynamic Memory Networks for Natural Language
  Processing}.
\newblock In \emph{ICML}.

\bibitem[{Li et~al.(2017)Li, Lam, Bing, and Wang}]{Li:2017:EMNLP}
Piji Li, Wai Lam, Lidong Bing, and Zihao Wang. 2017.
\newblock {Deep Recurrent Generative Decoder for Abstractive Text
  Summarization}.
\newblock In \emph{EMNLP}.

\bibitem[{Lin(2004)}]{Lin:2004:TSBO}
Chin-Yew Lin. 2004.
\newblock {ROUGE: A Package for Automatic Evaluation of Summaries}.
\newblock In \emph{TSBO}.

\bibitem[{Liu et~al.(2018)Liu, Saleh, Pot, Goodrich, Sepassi, Kaiser, and
  Shazeer}]{Liu:2018:ICLR}
Peter~J Liu, Mohammad Saleh, Etienne Pot, Ben Goodrich, Ryan Sepassi, Lukasz
  Kaiser, and Noam Shazeer. 2018.
\newblock {Generating Wikipedia by Summarizing Long Sequences}.
\newblock In \emph{ICLR}.

\bibitem[{Miao and Blunsom(2016)}]{Miao:2016:EMNLP}
Yishu Miao and Phil Blunsom. 2016.
\newblock {Language as a Latent Variable: Discrete Generative Models for
  Sentence Compression}.
\newblock In \emph{EMNLP}.

\bibitem[{Miller et~al.(2016)Miller, Fisch, Dodge, Karimi, Bordes, and
  Weston}]{Miller:2016:EMNLP}
Alexander Miller, Adam Fisch, Jesse Dodge, Amir-Hossein Karimi, Antoine Bordes,
  and Jason Weston. 2016.
\newblock {Key-Value Memory Networks for Directly Reading Documents}.
\newblock In \emph{EMNLP}.

\bibitem[{Na et~al.(2017)Na, Lee, Kim, and Kim}]{Na:2017:ICCV}
Seil Na, Sangho Lee, Jisung Kim, and Gunhee Kim. 2017.
\newblock {A Read-Write Memory Network for Movie Story Understanding}.
\newblock In \emph{ICCV}.

\bibitem[{Nallapati et~al.(2017)Nallapati, Zhai, and
  Zhou}]{Nallapati:2017:AAAI}
Ramesh Nallapati, Feifei Zhai, and Bowen Zhou. 2017.
\newblock {SummaRuNNer: A Recurrent Neural Network Based Sequence Model for
  Extractive Summarization of Documents}.
\newblock In \emph{AAAI}.

\bibitem[{Nallapati et~al.(2016)Nallapati, Zhou, dos Santos, Gulcehre, and
  Xiang}]{Nallapati:2016:CoNLL}
Ramesh Nallapati, Bowen Zhou, Cicero dos Santos, Caglar Gulcehre, and Bing
  Xiang. 2016.
\newblock {Abstractive Text Summarization Using Sequence-to-sequence RNNs and
  Beyond}.
\newblock In \emph{CoNLL}.

\bibitem[{Napoles et~al.(2012)Napoles, Gormley, and
  Van~Durme}]{Napoles:2012:NAACL-HLT}
Courtney Napoles, Matthew Gormley, and Benjamin Van~Durme. 2012.
\newblock {Annotated Gigaword}.
\newblock In \emph{NAACL-HLT AKBC-WEKEX}.

\bibitem[{Narayan et~al.(2018{\natexlab{a}})Narayan, Cohen, and
  Lapata}]{Narayan:2018:EMNLP}
Shashi Narayan, Shay~B. Cohen, and Mirella Lapata. 2018{\natexlab{a}}.
\newblock {Don't Give Me the Details, Just the Summary! Topic-Aware
  Convolutional Neural Networks for Extreme Summarization}.
\newblock In \emph{EMNLP}.

\bibitem[{Narayan et~al.(2018{\natexlab{b}})Narayan, Cohen, and
  Lapata}]{Narayan:2018:NAACL-HLT}
Shashi Narayan, Shay~B Cohen, and Mirella Lapata. 2018{\natexlab{b}}.
\newblock {Ranking Sentences for Extractive Summarization with Reinforcement
  Learning}.
\newblock In \emph{NAACL-HLT}.

\bibitem[{Oord et~al.(2016{\natexlab{a}})Oord, Dieleman, Zen, Simonyan,
  Vinyals, Graves, Kalchbrenner, Senior, and Kavukcuoglu}]{Oord:2016:SSW}
Aaron van~den Oord, Sander Dieleman, Heiga Zen, Karen Simonyan, Oriol Vinyals,
  Alex Graves, Nal Kalchbrenner, Andrew Senior, and Koray Kavukcuoglu.
  2016{\natexlab{a}}.
\newblock {WaveNet: A Generative Model for Raw Audio}.
\newblock In \emph{SSW}.

\bibitem[{Oord et~al.(2016{\natexlab{b}})Oord, Kalchbrenner, Espeholt, Vinyals,
  Graves et~al.}]{Oord:2016:NIPS}
Aaron van~den Oord, Nal Kalchbrenner, Lasse Espeholt, Oriol Vinyals, Alex
  Graves, et~al. 2016{\natexlab{b}}.
\newblock {Conditional Image Generation With Pixelcnn Decoders}.
\newblock In \emph{NIPS}.

\bibitem[{Over et~al.(2007)Over, Dang, and Harman}]{Over:2007:IPM}
Paul Over, Hoa Dang, and Donna Harman. 2007.
\newblock {DUC in Context}.
\newblock In \emph{IPM}.

\bibitem[{Park et~al.(2017)Park, Kim, and Kim}]{Park:2017:CVPR}
Cesc~Chunseong Park, Byeongchang Kim, and Gunhee Kim. 2017.
\newblock {Attend to You: Personalized Image Captioning with Context Sequence
  Memory Networks}.
\newblock In \emph{CVPR}.

\bibitem[{Pasunuru and Bansal(2018)}]{Pasunuru:2018:NAACL-HLT}
Ramakanth Pasunuru and Mohit Bansal. 2018.
\newblock {Multi-Reward Reinforced Summarization with Saliency and Entailment}.
\newblock In \emph{NAACL-HLT}.

\bibitem[{Paulus et~al.(2018)Paulus, Xiong, and Socher}]{Paulus:2018:ICLR}
Romain Paulus, Caiming Xiong, and Richard Socher. 2018.
\newblock {A Deep Reinforced Model for Abstractive Summarization}.
\newblock In \emph{ICLR}.

\bibitem[{Pereyra et~al.(2017)Pereyra, Tucker, Chorowski, Kaiser, and
  Hinton}]{Pereyra:2017:ICLR}
Gabriel Pereyra, George Tucker, Jan Chorowski, {\L}ukasz Kaiser, and Geoffrey
  Hinton. 2017.
\newblock {Regularizing Neural Networks by Penalizing Confident Output
  Distributions}.
\newblock In \emph{ICLR}.

\bibitem[{Rush et~al.(2015)Rush, Chopra, and Weston}]{Rush:2015:EMNLP}
Alexander~M Rush, Sumit Chopra, and Jason Weston. 2015.
\newblock {A Neural Attention Model for Abstractive Sentence Summarization}.
\newblock In \emph{EMNLP}.

\bibitem[{Salimans and Kingma(2016)}]{Salimans:2016:NIPS}
Tim Salimans and Diederik~P Kingma. 2016.
\newblock {Weight Normalization: A Simple Reparameterization to Accelerate
  Training of Deep Neural Networks}.
\newblock In \emph{NIPS}.

\bibitem[{Sandhaus(2008)}]{Sandhaus:2008:LDC}
Evan Sandhaus. 2008.
\newblock {New York Times Annotated Corpus}.
\newblock In \emph{LDC}.

\bibitem[{See et~al.(2017)See, Liu, and Manning}]{See:2017:ACL}
Abigail See, Peter~J Liu, and Christopher~D Manning. 2017.
\newblock {Get to the Point: Summarization with Pointer-Generator Networks}.
\newblock In \emph{ACL}.

\bibitem[{Singh et~al.(2017)Singh, Gupta, and Varma}]{Singh:2017:CIKM}
Abhishek~Kumar Singh, Manish Gupta, and Vasudeva Varma. 2017.
\newblock {Hybrid MemNet for Extractive Summarization}.
\newblock In \emph{CIKM}.

\bibitem[{Sukhbaatar et~al.(2015)Sukhbaatar, Weston, Fergus
  et~al.}]{Sukhbaatar:2015:NIPS}
Sainbayar Sukhbaatar, Jason Weston, Rob Fergus, et~al. 2015.
\newblock {End-to-end Memory Networks}.
\newblock In \emph{NIPS}.

\bibitem[{Sutskever et~al.(2014)Sutskever, Vinyals, and
  Le}]{Sutskever:2014:NIPS}
Ilya Sutskever, Oriol Vinyals, and Quoc~V Le. 2014.
\newblock {Sequence to Sequence Learning with Neural Networks}.
\newblock In \emph{NIPS}.

\bibitem[{Tan et~al.(2017)Tan, Wan, and Xiao}]{Tan:2017:ACL}
Jiwei Tan, Xiaojun Wan, and Jianguo Xiao. 2017.
\newblock {Abstractive Document Summarization with a Graph-based Attentional
  Neural Model}.
\newblock In \emph{ACL}.

\bibitem[{Vaswani et~al.(2017)Vaswani, Shazeer, Parmar, Uszkoreit, Jones,
  Gomez, Kaiser, and Polosukhin}]{Vaswani:2017:NIPS}
Ashish Vaswani, Noam Shazeer, Niki Parmar, Jakob Uszkoreit, Llion Jones,
  Aidan~N Gomez, {\L}ukasz Kaiser, and Illia Polosukhin. 2017.
\newblock {Attention is All You Need}.
\newblock In \emph{NIPS}.

\bibitem[{Wang and Ling(2016)}]{Wang:2016:NAACL-HLT}
Lu~Wang and Wang Ling. 2016.
\newblock {Neural Network-Based Abstract Generation for Opinions and
  Arguments}.
\newblock In \emph{NAACL-HLT}.

\bibitem[{Weston et~al.(2014)Weston, Chopra, and Bordes}]{Weston:2014:ICLR}
Jason Weston, Sumit Chopra, and Antoine Bordes. 2014.
\newblock {Memory Networks}.
\newblock In \emph{ICLR}.

\bibitem[{Yoo et~al.(2019)Yoo, Bahng, Chung, Lee, Chang, and
  Choo}]{Yoo:2019:CVPR}
Seungjoo Yoo, Hyojin Bahng, Sunghyo Chung, Junsoo Lee, Jaehyuk Chang, and
  Jaegul Choo. 2019.
\newblock {Coloring with Limited Data: Few-shot Colorization via
  Memory-Augmented Networks}.
\newblock In \emph{CVPR}.

\bibitem[{Yu and Koltun(2016)}]{Yu:2016:ICLR}
Fisher Yu and Vladlen Koltun. 2016.
\newblock {Multi-scale Context Aggregation by Dilated Convolutions}.
\newblock In \emph{ICLR}.

\bibitem[{Zhou et~al.(2017)Zhou, Yang, Wei, and Zhou}]{Zhou:2017:ACL}
Qingyu Zhou, Nan Yang, Furu Wei, and Ming Zhou. 2017.
\newblock {Selective Encoding for Abstractive Sentence Summarization}.
\newblock In \emph{ACL}.

\end{thebibliography}
\bibliographystyle{acl_natbib}

\clearpage
\appendix

\section{Implementation Details}
\label{sec:model_hyperparameters}

All the parameters are initialized with the Xavier method \cite{Glorot:2010:AISTATS}.
We apply the Adam optimizer \cite{Kingma:2015:ICLR} with $\beta_1 = 0.9, \beta_2 = 0.999$ and $\epsilon = 1e-8$.
We apply weight normalization \cite{Salimans:2016:NIPS} to all layers.
We set learning rate to 0.001 and clip gradient at 0.3.
At every 4 epochs, we divide learning rate by 10 until it reaches 0.0001.
We train our models up to 12 epochs for TIFU-short and 60 epochs for TIFU-long.

Table \ref{tab:hyperparams} summarizes the setting of hyperparameters for our model in all experiments on TIFU-short/long dataset, Newsroom abstractive subset and XSum.

\section{Novel N-gram Ratios}
\label{sec:novel_ngram_ratio}

Table \ref{tab:dataset-ngram} compares the ratios of novel N-grams in the reference summary between datasets.
Following \citep{See:2017:ACL, Narayan:2018:EMNLP}, we compute this ratio as follows;
we first count the number of N-grams in the reference summary that do not appear in the source text and divide it with the total number of N-grams.
The higher the ratio is, the less the identical N-grams are in the source text.
The \texttt{CNN/DailyMail}, \texttt{New York Times}, \texttt{Newsroom} datasets all, for example, exhibit low novel 1-gram ratios as 10.3\%, 11.0\%, 15.6 \% respectively.
This means that about 90\% of the words in reference summary already exist inside the source text.
It is due to that the summaries from formal documents (\eg news and academic papers) tend to have same expressions with the source documents.
Therefore, these datasets may be more suitable for extractive summarization than abstractive one;
on the other hand, our dataset is more abstractive.

We also compare the novel N-gram ratio for \texttt{XSum} and three subsets of \texttt{Newsroom};
(i) \texttt{Newsroom-Ext}, a subset favorable for extractive methods,
(ii) \texttt{Newsroom-Mix}, a subset favorable for mixed methods, and
(iii) \texttt{Newsroom-Abs}, a subset favorable for abstractive methods.
We summarize two interesting observations as follows.
First, as expected, the more favorable for abstractive methods is, the higher novel n-gram ratio is.
Second, novel n-gram ratios of \texttt{Newsroom-Abs} and \texttt{XSum} are higher than those of our dataset, even though their data sources are news publications.
Thus, we argue that novel n-gram ratios are pretty good but not a sufficient measure to find extractive bias in the summarization dataset.

\begin{table}[t]
    \centering
    \small
    \setlength{\tabcolsep}{2pt}
    \begin{tabular}{|c|c|c|}
        \hline
        \textbf{Description}      & \multicolumn{2}{c|}{\textbf{Common Configurations}} \\
        \hline
        Initial learning rate     & \multicolumn{2}{c|}{0.001} \\
        Embedding dimension ($d^{emb}$)       & \multicolumn{2}{c|}{300} \\
        Kernel size ($k$)               & \multicolumn{2}{c|}{3} \\
        Dilation rate ($d$)             & \multicolumn{2}{c|}{$2^l$} \\
        \hline
        \hline
        \textbf{Description}      & \textbf{TIFU-short} & \textbf{TIFU-long} \\
        \hline
        Grad clip                 & 0.3 & 0.3 \\
        \# of encoder layers      & 9 & 8 \\
        \# of decoder layers      & 3 & 5 \\
        Layers used for memory ($\mathbf m$)    & \{3, 6, 9\} & \{4, 8\}\\
        Smoothing parameter ($\epsilon$)       & 0.1 & 0.05 \\
        \hline
        \textbf{Description}      & \textbf{Newsroom-Abs} & \textbf{XSum} \\
        \hline
        Grad clip                 & 0.3 & 0.8 \\
        \# of encoder layers      & 10 & 9 \\
        \# of decoder layers      & 6 & 6 \\
        Layers used for memory ($\mathbf m$)    & \{3, 6, 10\} & \{4, 7, 9\} \\
        Smoothing parameter ($\epsilon$)       & 0.05 & 0.05 \\
        \hline
    \end{tabular}
    \caption{
        Model hyperparameters in experiments on TIFU-short/long, Newsroom abstractive subset and XSum.
    }
    \label{tab:hyperparams}
\end{table}

\begin{table}[t]
    \centering
    \small
    \setlength{\tabcolsep}{3pt}
    \begin{tabular}{|c|cccc|}
        \hline
                                    &\multicolumn{4}{c|}{\textbf{Novel N-gram Ratio}}  \\
        \hline
        Dataset                     & 1-gram    & 2-gram    & 3-gram    & 4-gram        \\
        \hline
        {\tt CNN/DailyMail}         & 10.3      & 49.9      & 70.5      & 80.3          \\
        {\tt NY Times}              & 11.0      & 45.5      & 67.2      & 77.9          \\
        {\tt Newsroom}              & 15.6      & 45.4      & 57.2      & 62.2          \\
        {\tt Newsroom-Ext}          & 1.5       & 5.9       & 8.9       & 11.1          \\
        {\tt Newsroom-Mix}          & 11.6      & 47.0      & 66.5      & 76.8          \\
        \hline
        {\tt Newsroom-Abs}          & 33.9      & 83.9      & 97.1      & 99.5          \\
        {\tt XSum}                  & 35.8      & 83.5      & 95.5      & 98.5          \\
        \hline
        {\tt TIFU-short}            & 29.7      & 71.5      & 88.1      & 93.8          \\
        {\tt TIFU-long }            & 27.4      & 76.7      & 92.5      & 97.0          \\
        \hline
    \end{tabular}
    \caption{
        Comparison of novel N-gram ratios between Reddit TIFU and other summarization datasets.
    }
    \label{tab:dataset-ngram}
\end{table}

\section{More Examples}
\label{sec:examples}

\begin{figure*}[t] \begin{center}
    \includegraphics[width=\linewidth]{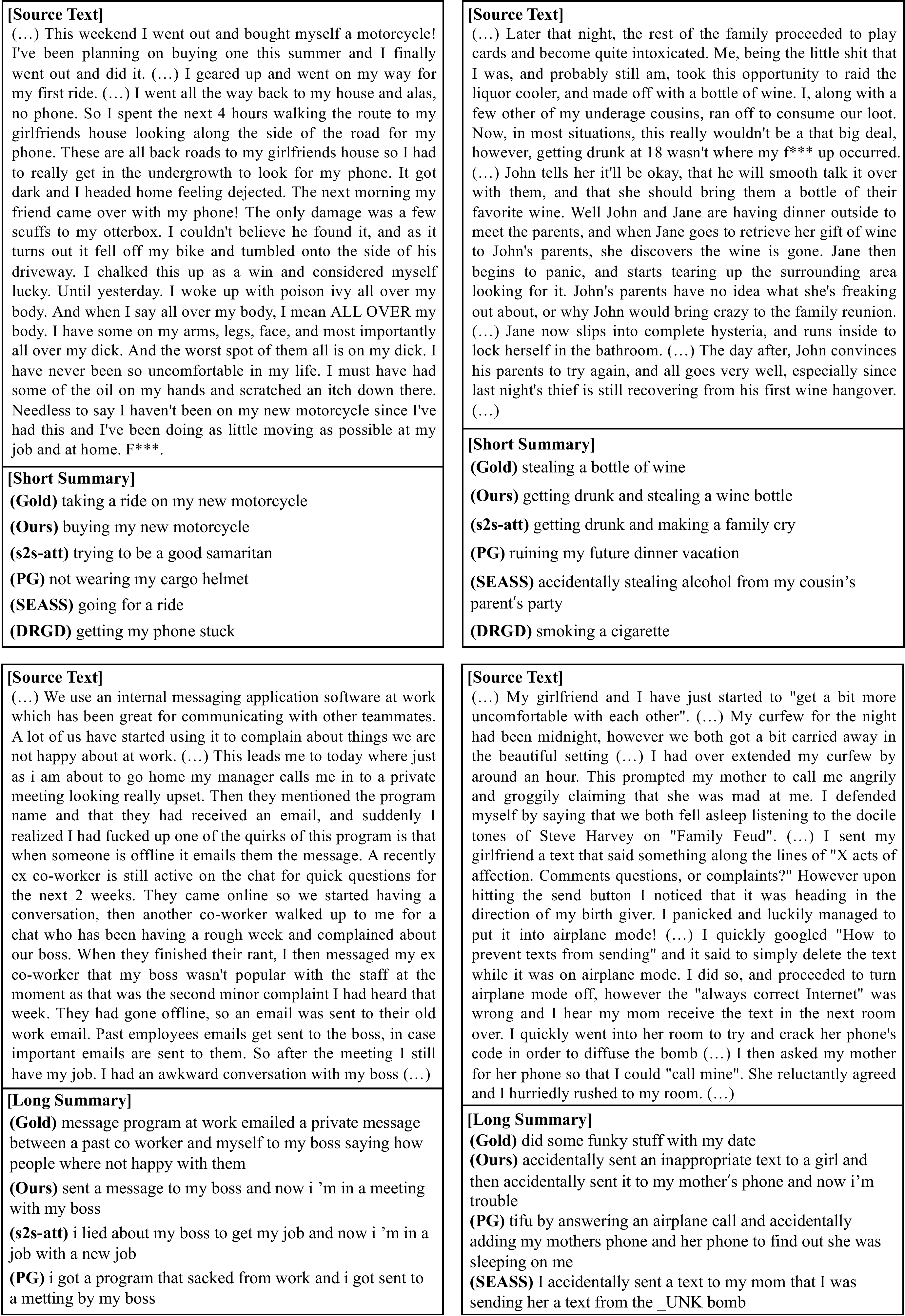}
    \caption{Examples of abstractive summary generated by our model and baselines. In each set, we too show the source text and reference summary. 
    }
    \label{fig:examples}
\end{center} \end{figure*}

Figure \ref{fig:examples} illustrates selected examples of summary generation.
In each set, we show a source text, a reference summary and generated summaries by our method and baselines.
In the examples, while baselines generate summary by mostly focusing on some keywords, our model produces summary considering both keywords and the whole context thanks to the multi-level memory.

\end{document}